\documentclass{article}
\usepackage{spconf,amsmath,graphicx}
\usepackage{graphicx}
\usepackage{setspace}

\usepackage[round]{natbib}

\title{Deep Learning based Multi-Label Image Classification of Protest Activities}
\name{Yingzhou Lu, Kosaku Sato, Jialu Wang}
\address{ Electrical and Computer Engineering Department\\ Virginia Tech\\ Arlington,VA
 22203, USA\\ 
 George Washington University\\
 Washington, DC 20052, USA\\
 Email: lyz66@vt.edu, Ksato@vt.edu,
 jialu@gwu.edu
 }
 
\begin{document}
 
 \maketitle
\setlength{\labelsep}{0.3em} 
\begin{spacing}{0.9}
\begin{abstract}
With the rise of internet technology amidst increasing urbanization rates, sharing information has never been easier, thanks to globally-adopted platforms for digital communication. The resulting output of massive amounts of user-generated data can be used to enhance our understanding of significant societal issues, particularly for urbanizing areas. In order to better analyze protest behavior, we enhanced the GSR dataset and manually labeled all the images. We used deep learning techniques to analyze social media data to detect social unrest through image classification, which performed well in predicting multi-attributes. Then, we used map visualization to display protest behaviors across the country.
\end{abstract}
\setlength{\parsep}{-0.8ex} 
\vspace{-0.5em}

\section{Introduction}
\vspace{-0.5em}
\label{sec:intro}
The study of protest activities plays a profound role in sociologists and scholars' studying citizens’ political behavior. With the advancement of social media networks, people now share an unprecedented amount of user-generated content in the form of text, images, and videos on the web. Classification of social media data not only helps in understanding online behavior, but also elucidates significant priorities of urban populations that carry real-life consequences. Using social media data, we focuses on social unrest in the form of public protest images, specifically for Latin American countries. 

The traditional approach to the study of social media dataset focused on using natural language processing to monitor how hashtags and links are used by the user and the propagation of those items to other users. However, these approaches may not effectively capture some important features or details of protest activities. For instance, we may be interested in knowing details such as whether there was a large crowd involved in the protest, if polices were present, or what the demographics (young or adults) of protesters carrying a sign. Our approach uses image processing to capture those features of the protest activities\cite{fu2021probabilistic}.

We took several approaches in image classification of social media data: our initial approach is to utilize traditional machine learning methods such as Support Vector Machine (SVM)\cite{weston1999support} and a deep learning method like Convolutional Neural Networks (CNNs)\cite{krizhevsky2012imagenet} which have shown some advantages in large-scale image and video analysis. Traditional machine learning method, such as SVM, can be used to classify images with good accuracy; however, as the volume of data and number of classes for recognition increases, the deep learning approaches becomes the more advanced approach for object recognition.

\section{Literature Review}
As our objective of our model is to detect protest activities using the image, the preliminary work relevant to our study is the EMBERS system by Naren, Patrick and et el\cite{ramakrishnan2014beating}. The EMBERS system continuously monitor the social media dataset such as Twitter, Facebook, news pages, and use data mining to process the trend to predict the protest activities in South America regions. Their planned protest model based on custom multi-lingual lexicon matching predicted the protest activities with precision and recall rate of 0.69 and 0.82 respectively. However, their approach does not capture the additional features of the protests such as demography.

Moreover, image classification is classical topic in computer vision area which aims to predict and assign each given image a specific label from several categories. However, background clutter, occlusion and variation in image scale make the computer vision tasks more challenging. The traditional approaches to perform image classification includes k-nearest neighbor and SVM algorithms\cite{yi2018enhance}. k-nearest neighbor is one of the simplest classification algorithm that aims at labeling an image based on the best fit result, but the model is usually not robust to noise or imbalanced class dataset\cite{zhang2006svm}. Similarly, SVM, which is originally proposed as a binary classification by Cortes and Vapnik\cite{cortes1995support} is another classical approach to perform classification and it has shown a better performance than the k-nearest neighbor in some applications\cite{boiman2008defense}. Furthermore, our approach adopts a deep learning based on CNNs based on the study of the visual cortex of the human brain which have shown a great success recently in many computer vision applications\cite{aghdamguide}. 

\begin{figure}
\centering
\includegraphics[width=80mm]{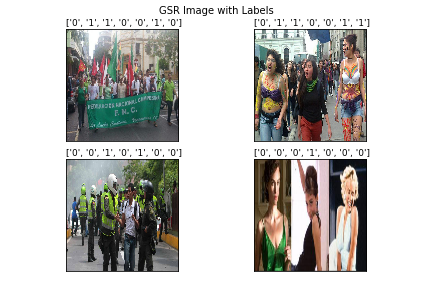}
\caption{Labeled GSR images in GSR dataset}
\label{GSR images}
\end{figure}

\section{OSI Dataset}
The OSI (Open Source Indicators) dataset, provided by computer science department from Virginia Tech, is MITRE's gold standard report (GSR) of protests organized by surveying newspapers for civil unrest reports. The dataset also contains large samples of non-protest images that were collected in the process. There are 48,713 images in GSR and 40,647 non-protest images. High-confidence images indicate a top image among multiple images embedded in the articles. High-confidence images within datasets indicate a top image among multiple images
embedded in the articles\cite{jayachandra2020besober}. High-confidence GSR images relevant to GSR articles based on social protest total 7,884, and low-confidence GSR images total 40,829.
\subsection{Details of the Image Labels}
Table 1 shows the visual attributes that characterize the protests which we used to label each image from the GSR dataset. Out of 40,647, a total of 9,504 images were hand-picked to train and test our prediction models by excluding bad data points that are obviously irrelevant to the social activities that we are interested in detecting\cite{yang2020coordinating}. The Annotated images consist of 327 fire, 1,943 flag, 7,347 large crowd, 248 other, 2,159 police, 4,462 sign, and 1,233 student images. Fig. 1 contains sample images with their class labels. Each image has a label with vector of length 7 that has "0"s and "1"s corresponding to the index number of visual attributes. The "Other" label was inferred from the absence of positive class labels across categories.

\subsection{Challenges of the Dataset}
\begin{figure}
\centering
\includegraphics[width=80mm]{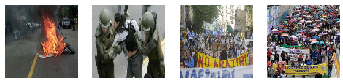}
\caption{Sample protest images from training set}
\label{Sample protest}

\centering
\includegraphics[width=80mm]{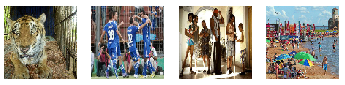}
\caption{Sample non-protest images from training set}
\label{Sample non-protest}
\vspace{-0.7em}
\end{figure}

There are a few inherent challenges in our training dataset. First, some attributes of protest are commonly shared with non-protest images. For instance, Figure \ref{Sample protest} shows a sample of protest images used to train the machine learning and deep learning models. Images with a protest class label often depicted fire, police, handwritten signs, and large crowds. However, Figure \ref{Sample non-protest} shows a sample of non-protest images in which large crowds are also frequently seen attribute while it also comprised of a variety of other objects (such as animals or soccer players). Second, we have imbalanced dataset where it does not have exactly equal number of instances in each class. This issue is mainly defined by the specific subject or attribute we set up in our problem. The training of the classifiers in imbalanced dataset can cause the trained model classifying images as images of majority class most of the times and under-represent the minority class. 

\begin{table}[]
\centering
\caption{Visual attributes of protest images}
\resizebox{0.75\textwidth}{!}{\begin{minipage}{\textwidth}
\label{graph}
\begin{tabular}{|c|c|c|c|}
\hline
  & Class label & Description   & Sample size                                       \\ \hline
0 & Fire        & Presence of active fire  & 327                            \\ \hline
1 & Flag        & Presence of flag       & 1943                              \\ \hline
2 & Large Crowd & Presence of roughly more than 20 people  & 7347            \\ \hline
3 & Other       & None of the above or the below    & 248                   \\ \hline
4 & Police      & Presence of police &2159                                 \\ \hline
5 & Sign        & Presence of a protest sign &  4462 \\ \hline
6 & Student     & Presence of young students  & 1233                         \\ \hline
\end{tabular}%
\end{minipage}}
\vspace{-1.7em}
\end{table}

\begin{table}[]
\centering
\caption{Increased sample size after image augmentation}
\resizebox{0.75\textwidth}{!}{\begin{minipage}{\textwidth}
\label{graph}
\begin{tabular}{|c|c|c|}
\hline
  Class label & Image transformation used   & New sample size                                       \\ \hline
Fire        &  
\vtop{\hbox{\strut Flipping, Scaling, Translation, Noise,}\hbox{\strut  Affine Transform, Perspective Transform,}\hbox{\strut  Intensity, Contrast, Filters, Crop, Shear}} & 4,578                           \\ \hline
Flag        & Flipping, Noise, Affine Transform     & 5,829                             \\ \hline
Large Crowd & --  & 7347            \\ \hline
Other       & \vtop{\hbox{\strut Flipping, Scaling, Translation, Noise,}\hbox{\strut  Affine Transform, Perspective Transform,}\hbox{\strut  Intensity, Contrast, Filters, Crop, Shear}} & 3,472                   \\ \hline
Police      & Affine Transform, Noise &6,477                                \\ \hline
Sign        & -- &  4462 \\ \hline
Student     & Flipping, Noise, Affine Transform, Crop  & 6,165                         \\ \hline
\end{tabular}%
\end{minipage}}
\vspace{-1.7em}
\end{table}

\begin{figure}
\centering
\includegraphics[width=80mm]{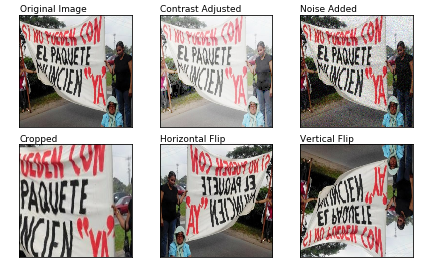}
\caption{Example images from the dataset after transformation}
\label{Image Augmentation}
\end{figure}
\subsection{Image Augmentation for Imbalanced Dataset}
One way to balance the imbalanced classes is to use image augmentation. Data augmentation is a method to artificially increase sample size of the training images through various pre-processing or combinations of multiple pre-processing of the image such as adding noise, flipping and re-scaling. Table 2 summarizes the different techniques we adopted to perform image augmentation for minority classes. Image augmentation has been considered as a promising method to improve the performance of prediction model. For instance, adding noise to our observation can help make the prediction model more robust in the face of social media dataset and prevents it from overfitting \cite{perez2017effectiveness}. Moreover, Python's scikit-image library has the full list of the available image transformation but we have adopted the thirteen of them. Those are horizontal and vertical flipping, affine transfomration, perspective transformation, rescaling, cropping, blurring, changing contrast and intensity, gaussian and exposure filter, translation, and shearing. Fig. 4 shows the example of such transformations of the images. As a result of implementing image augmentation, we were able to significantly improve the sample size of the minority classes; for instance, the sample size of 'fire' class increased from 327 to 4,578 as seen in Table 2. Also, as an alternative approach, we considered oversampling using Synthetic Minority Over-sampling Technique (SMOTE) \cite{chawla2002smote}. However, we believe that image augmentation can create more variation in training images to prevent over fitting, and hence we did not utilize SMOTE in this paper. 

\section{Multi-Label Image Classification}
Multi-label learning is a form of supervised learning where the
classification algorithm learns from a set of images in which an
image can belong to one or more classes. The goal of multi-label
image classification is to predict a set of class labels for the input
image. A more generalized approach is multi-class learning where
each image is limited to one correct class label. Multi-label classification
and prediction is more practical since the many real world
problems involve multiple objects belonging to different categories.
Multi-label classification is also applicable various domains such as
text, video, and scene classification. For a typical multi-label
image, objects of different categories in each image are located at
varying positions with differing scale, zoom, size, and pose\cite{won2017protest}. For
example, two images labeled as 'police' and 'fire' may have different
spatial arrangements of identified objects. Although factors such as
differing arrangements or occlusion can contribute to the inaccuracy
of multi-label classification, we expect reasonable results with
a sufficiently large dataset. Details of implementation of multi-label classification for SVM and CNNs will be discussed in the following sections.

\section{Approach of Image Classification}

In our approach, we utilized multi-label SVM and CNNs to detect protest attributes in image classification. First, multi-label SVM will be explained in details. In our proposal, SVM is a baseline model to evaluate the performance of the prediction model using CNNs. As the problem requires a large dataset for training and good accuracy of classifying many protest attributes, we believe that CNNs will perform better than SVM.

\subsection{Support Vector Machine}
SVM is originally proposed as a binary classification by Cortes and Vapnik\cite{cortes1995support}, but the model has been extended to apply to multi-label classification problems\cite{weston1998multi}. 

\textbf{One-vs.-All:} One-vs.-All is a classical approach to solve k-class pattern recognition problem. It involves training a single binary classifier per class, with the samples of one class as positive samples while other samples are set as negative. More specifically, using this method, n-th classifier finds a hyperplane between class $n$ and the rest of the classes\cite{weston1998multi}. A point where the distance from the margin is maximal is assigned to the class. We aim at detecting seven classes so that this strategy requires the training of seven different SVMs. During testing the models, all classifiers would vote 'true' by predicting that a testing sample belongs to their class. In the end of testing, a sample is classified by the ensemble as the class that has the highest number of votes. One-vs.-All is widely used in multi-label classification.

\textbf{Weighting Hyper-Parameter for SVM:}
Imbalanced dataset causes misclassification of images that belonging to the minority class impacted more heavily than that of the majority class because the frequency of the minority class is rare compared to that of the majority class. In order to mitigate over fitting of training classifier resulted from the imbalanced data, we propose the modification of hyperparameter $C$ in SVM's objective function which determines the penalty for misclassifying the objects. Instead of defaulting $C$ to be one, $C_k$ belonging to class $k$ will have different values as shown in (4). 
\begin{equation}
\setlength{\abovedisplayskip}{1pt} 
\setlength{\belowdisplayskip}{1pt}
\begin{aligned}
C_k  = C\cdot \frac{n}{kn_j}\\
\end{aligned}
\end{equation}  
As you can see, the updated $C_k$ value will be inversely proportional to instances of $j$ class in order to increase $C_k$ value for the minority class in order to mitigate under-representation issue. $k$ is the number of class and $j$ is the sample size belonging to the class.

 \subsection{Convolutional Neural Networks}
\subsubsection{Architecture}
CNNs consist of input, convolution, activation function, pooling,
deep layers, and output layers. Throughout
the training of the network, the parameters are updated except
for the ones between convolution and pooling. There are some
important properties of the convolution layer. Some patterns are
smaller than the entire image so that the image can be subsampled
to reduce the image size; this is to train fewer parameters in the
neural network. The same patterns can appear in different regions
so that the same set of parameters can be used to reduce computation. 

$\textbf{Convolutional Layers:}$
Convolution of a filter on an input image is a point-wise multiplication operation. The activation function, which in our case is Rectified Linear Unit (ReLU) activation, is applied on each image separately in an element-wise fashion to create activation maps based on outputs of the convolution \cite{aghdamguide}.

$\textbf{Activation Function:}$ Nair and Hinton introduced the non-saturating nonlinearity $f(x)=\max(0,x)$, also known as the ReLU, which has gained popularity in the deep-learning community because of its fast computing time \cite{krizhevsky2012imagenet}.Hence, our model applied the ReLU nonlinearity function to the output of every convolutional and fully-connected\cite{krizhevsky2012imagenet}.

$\textbf{Pooling Layer:}$
In our model, we applied a 2x2 filter size with a 2-length stride after each layer with the option of max-pooling. Max-pooling applies the filters and the stride to the input and returns the maximum value, dropping the non-max values in each sub-region that convolution is applied. 

$\textbf{Fully Connected Layers:}$
The fully connected layer uses those inputs to produce N-dimensional vectors, where N is the number of classes needed for prediction.  

$\textbf{Loss Function:}$
In our model, we used sigmoid cross-entropy for multi-label classification. Cross-entropy is used to define the loss function in training the network in which the model is penalized if it estimates a low probability for the target class \cite{nielsen2015neural}. 
\begin{equation}
\setlength{\abovedisplayskip}{1pt} 
\setlength{\belowdisplayskip}{1pt}
\begin{aligned}
J(\theta)&=-\frac{1}{m} \sum_{i=1}^{m} \sum_{k=1}^{K}[y_k^i\log(\hat{p}_k^i)]
\end{aligned}
\end{equation}

For the loss function, we used Adaptive Moment Estimation (ADAM) \cite{kingma2014adam}. ADAM keeps track of a learning rate for each network weight and computes individual adaptive learning rates for different parameters based on estimates of first and second moments of the gradients \cite{kingma2014adam}. Since ADAM is an adaptive rate learning algorithm, it requires less tuning. The default learning rate 0.001 is often used to support usability of the algorithm.

Table 3 shows our CNNs architecture. There are three convolutional layers with ReLU, and max-pooling is used to down-sample the image after each convolutional layer. The filter size for max-pooling is $2\times2$ so that output image of the max-pooling is half the size of the input. The convolutional layers uses the filter size of $3\times3$ while length of stride equal to 2. The first fully connected layer (FC1) is a vector with a length of 1024, and the second fully connected layer (FC2) is a vector of length 7 which is the number of class labels to predict visual attributes of protest images.

\begin{table}[]
\centering
\caption{Architecture of Multi-Label Classification CNN}
\resizebox{0.68\textwidth}{!}{\begin{minipage}{\textwidth}
\label{graph}
\begin{tabular}{|c|c|c|c|c|c|c|}
\hline
Layer & Feature Map & Feature Size & Filter Size & Stride & Pad & Activation \\ \hline
FC2         & -      & 7       & -   & - & -    & Sigmoid \\ \hline
FC1         & -      & 1024    & -   & - & -    & ReLU    \\ \hline
Max-Pooling & 128    & 4x4     & 2x2 & 2 & Same & -       \\ \hline
Conv3       & 128    & 7x7     & 3x3 & 2 & Same & ReLU    \\ \hline
Max-Pooling & 64     & 14x14   & 2x2 & 2 & Same & -       \\ \hline
Conv2       & 64     & 28x28   & 3x3 & 2 & Same & ReLU    \\ \hline
Max-Pooling & 32     & 56x56   & 2x2 & 2 & Same & -       \\ \hline
Conv1       & 32     & 112x112 & 3x3 & 2 & Same & ReLU    \\ \hline
Input       & 3(RGB) & 224x224 & -   & - & -    & -       \\ \hline
\end{tabular}%
\end{minipage}}
\vspace{-0.5em}
\end{table}
\vspace{-1.1em}
\subsubsection{Evaluation Method of Multi-label Classification}
Evaluation of multi-label classification has a notion of being partially correct. One way to evaluate the classification is label-set based accuracy or exact match that considers partially correct as incorrect. On the other hand, evaluation of label-based accuracy is carried out on a per label basis\cite{chen2021learnable}. The calculation method of label-set accuracy, where a predicted set of labels $\hat{y}$ must exactly match the ground truth $y$, is shown in equation (3) \cite{read2011classifier}; 0/1 loss dictates that any label vector not predicted perfectly will be given a zero score.
\begin{equation}
\setlength{\abovedisplayskip}{1pt} 
\setlength{\belowdisplayskip}{1pt}
 0/1\quad loss = 1 - \frac{1}{N} \sum_{i=1}^{N} 1_{y_i=\hat{y_i}}
\end{equation}
Label-based accuracy is more lenient approach to evaluate the performance since it does not consider multi-label problem as a whole. When each label has a separate binary evaluation, we have hamming loss which is shown in the following equation:
\begin{equation}
\setlength{\abovedisplayskip}{1.1pt} 
\setlength{\belowdisplayskip}{1pt}
Hamming \quad Loss = 1-\frac{1}{NL} \sum_{i=l}^{N} \sum_{j=l}^{L} 1_{y_i=\hat{y_i}}
\end{equation}
We adapted both approaches to evaluate the performance of our multi-label classifier.
\subsubsection{Threshold Selection}
The fully connected layer represents a vector containing probability for each class. The threshold function can be used to obtain a multi-label prediction $\hat{y}$. Specifically, we used the Matthews Correlation Coefficient (MCC) which is an evaluation metric of binary classification. The MCC is a correlation coefficient for ground truth versus predictions and varies between -1 and 1, where 1 represents a perfect prediction \cite{gorodkin2004comparing}. The MCC is given by the following equation (5).
\begin{equation}
MCC = \frac{T_p \times T_n - F_p \times F_n }{\sqrt{(T_p+F_p)(T_p+F_n)(T_n+F_p)(T_n+F_n)}}
\end{equation}

For multi-label classification, the MCC is defined in terms of a confusion Matrix C for K classes in the equation (6) .
\begin{equation}
MCC = \frac{c \times s - \sum_{k}^{K} p_kxt_k }{\sqrt{(s^2-\sum_{k}^{K}p_k^2)(s^2-\sum_{k}^{K}t_k^2)}}
\end{equation}

The values $t_k = \sum_{i}^{K}C_{ik}$  is the number of times class $k$ truly happened. $p_k = \sum_{i}^{K}C_{ki}$  is the number of times class $k$ was predicted\cite{tian2021explore}. $c = \sum_{i}^{K}C_{kk}$  is the total number of samples correctly predicted. $s = \sum_{i}^{K}\sum_{j}^{K}C_{ij}$  is the total number of samples.

\subsection{Advantages and Disadvantages of SVM and CNN}
In image classification, there are advantages and disadvantages for both SVM and CNNs. Theoretically, SVM is very good at finding the margin and hyperplane for classification, and it is very robust for high dimensional data\cite{zhang2006svm}. However, SVM model is sensitive to noise, for example, if there is a noise in background or a visible object in one image is occluded or partially blocked by scenes in other, it will have a negative impact on the performance of the classification model \cite{cortes1995support}. Moreover, since one-vs-all involves training a binary classifier for all classes, computation time can be very expensive.

On the other hand, the main advantage of CNNs is that it can be used to extract important image features with a sufficiently large dataset\cite{lecun2015deep}. The performance of CNNs classifier largely depend on the size of the dataset. The bottle necks in training the CNNs model are computational time and memory used to retain activation from forward pass and error gradients computation when dataset is very large\cite{sun2008efficient}. However, an efficient parallel computation with a help of GPU or training a model in mini batches can be used to mitigate those issues to some degrees. Also, there are many parameters for CNNs that need to be set by the users in order to train a robust and good prediction model. 

\section{Experiment}
\vspace{-0.5em}
We implemented K-class SVM and CNNs using scikit-learn and TensorFlow libraries in Python 3.5. In the experiment, we  used  our personal server desktop which runs on Windows with 2 Intel Xeon  E5-2630 V3 CPU 2.4 GHz and 8 small cores with RAM size of 64GB.  First, we split the sample images via image augmentation from Table 2 into training and testing (the ration of 80\% and 20\% respectively) and then we re-sized each image to 224x224 with 3 color channels. However, the implementation of SVM in scikit-learn does not adopt online learning so that we had to down-sample the sample images from 31,472 to 12,000 to avoid memory limit error. On the other hand, with a help of mini-batches, we used all of the data points without any down-sampling for CNNs model.

We trained a baseline SVM using One-vs.-All method. Our final setting of SVM consisted of max iterations to be ran as 4000 to ensure it converges. Also, we experimented weighting of hyperparameter in equation (1) but it did not improve the result so our final setting of the weight parameters for each classes are set to 1. For CNNs, we used the learning rate of 0.001 which is a standard rate and mini batch size of 202 images with numbers of batches as 125. 

\section{Results and Discussion}
\subsection{Evaluation}
Evaluation was conducted on each class. Since we have the manual label as ground truth, we calculated accuracy, the precision rate, recall rate, and F1 score for each class respectively. Specifically, the following is the equations we used to calculate each evaluation criteria: Precision rate$=\frac{TP}{TP+FP}$, Recall rate$=\frac{TP}{TP+FN}$, Accuracy$=\frac{TP+TN}{TP+TN+FP+FN}$. F1 score is the harmonic mean of precision rate and recall rate.

\begin{table}[h!]
\centering
\caption{Evaluation of SVM Model Prediction}
\resizebox{0.75\textwidth}{!}{\begin{minipage}{\textwidth}
\label{graph}
\begin{tabular}{|c|c|c|c|c|c|c|c|}
\hline
             & Fire & Flag & Large Crowd & Other & Police & Sign & Student \\ \hline
Accuracy (\%)   & 88 & 74  & 60         & 89   & 77    & 63   & 79     \\ \hline
Precision (\%) & 53   & 53   & 63          & 53    & 52     & 55   & 50      \\ \hline
Recall (\%) & 35   & 22   & 60          & 24    & 30     & 39   & 19      \\ \hline
F1 Score (\%) & 42   & 31   & 61          & 33    & 38     & 45   & 27      \\ \hline
\end{tabular}
\end{minipage}}
\end{table}

\begin{table}[]
\centering
\caption{Evaluation of CNNs Model Prediction}
\resizebox{0.75\textwidth}{!}{\begin{minipage}{\textwidth}
\label{graph}
\begin{tabular}{|c|c|c|c|c|c|c|c|}
\hline
             & Fire & Flag & Large Crowd & Other & Police & Sign & Student \\ \hline
Accuracy (\%)   & 91  & 72  & 71         & 89   & 76    & 61   & 76     \\ \hline
Precision (\%) & 76   & 48   & 74          & 52    & 45     & 51   & 32      \\ \hline
Recall (\%) & 67   & 36   & 73          & 37    & 32     & 46   & 23      \\ \hline
F1 Score (\%) & 71   & 41   & 73          & 43    & 38     & 49   & 27      \\ \hline
\end{tabular}
\end{minipage}}
\vspace{0.2em}
\end{table}

\begin{figure}[h!]
\centering
\vspace{-1.3em}
\includegraphics[width=90mm]{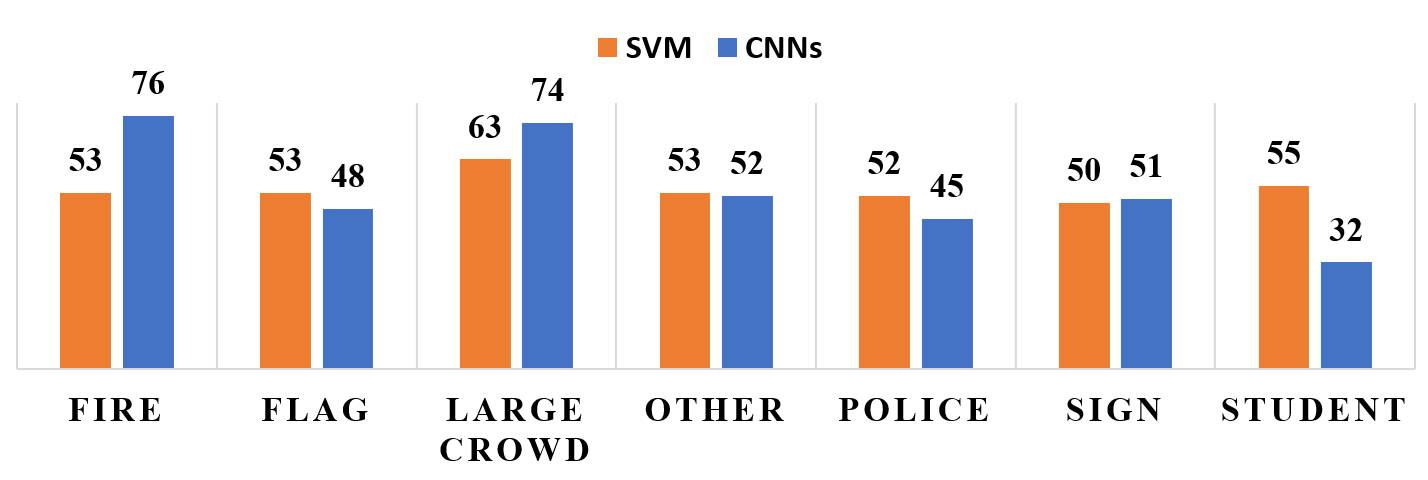}
\caption{SVM vs CNN Precision Rate}
\label{pic}
\end{figure}

\begin{figure}[h!]
\centering
\vspace{-1.3em}
\includegraphics[width=90mm]{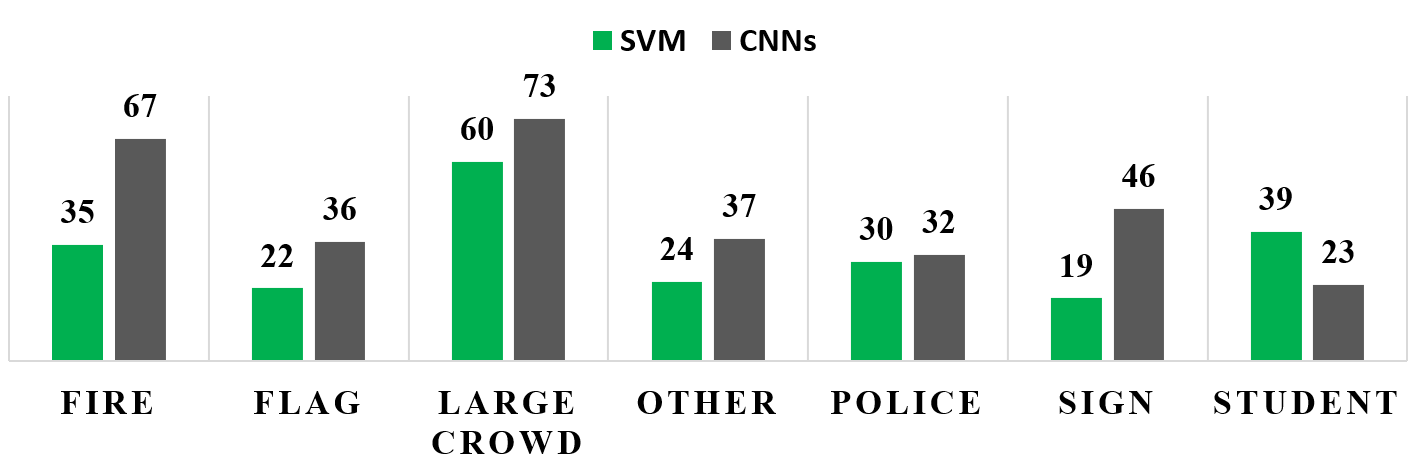}

\caption{SVM vs CNN Recall Rate}
\label{pic}
\end{figure}

\begin{figure}
\centering
\vspace{-1.3em}
\includegraphics[width=30mm]{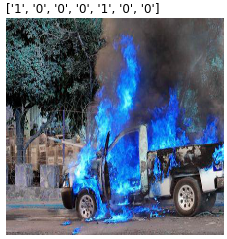}
\includegraphics[width=30mm]{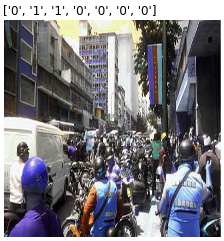}
\caption{Image of a burning vehicle with police in background (top); image of bikers and a flag (bottom)}
\label{pic}
\end{figure}

\subsection{Result}
Training SVM model with one-vs.-all method took longer than 12 hours and consistently consumed 70-90\% of available memory on our machine whereas the CNNs model only took less than half of the training time with much less consumption of memory with a help of mini-batch. Therefore, we were able to obtain the results by CNNs easier and faster than the SVM. Table 4 and 5 shows the accuracy, precison, recall, and F1 score of each predicted lables for SVM and CNNs respectively. We also plotted the precision and recall of the two models side by side in histogram in Fig. 5 and 6 to compare the performance of the two models. As you can see, their overall performance is comparable to each other but recall using CNNs is slightly better than that of SVM. 

For the evaluation of CNNs prediction, We calculated the best threshold using MCC to transform the probability of each label from the fully connected layer into the 7 predicted class labels: the calculated thresholds are 0.2, 0.4, 0.7, 0.6, 0.5, 0.4, 0.5 for 'fire', 'flag', 'large crowd', 'other', 'police', 'sign', and 'student' respectively. Prediction accuracy per label reached almost 77\% on average. Fig. 7 shows sample test images of 'fire' and 'police' on the left and 'large crowd' and 'police' on the right. Our CNNs model predicted them correctly but when we evaluated our classifier model with a large dataset, we found that our label-set accuracy was very low around 20\% due to the challenges of exact matching on a multi-label classifier.  

\subsection{Future work}
From the experiment, we learnt that we were able to get reasonable performance using both SVM and CNNs model to predict each class label separately but the bottom line performance of our prediction model is still not desirable: our goal is to increase the accuracy of exact matching. Therefore, Future work can be done in following aspects. First, we can apply state-of-the-art algorithm like Generative Adversarial Network to generate more training samples, which would be helpful to prevent over fitting. Second, we modify the equation for SVM to enhance the classifier, and improve the deep learning model\cite{yan2018multi}. Moreover, The main limitation of our image classification approach is that it does not consider the credibility of the source in decision making, and hence requires assessment of the social media source or of each image posted on the web. Also, there are privacy protection concern in using both social media and image data\cite{Chai2018}. In other further research, we will merge image and text data like article headlines and descriptions associated with each image which should help improve the performance of prediction model. We will conduct privacy protection procedure such as Randomized Response \cite{Chai2019} to the data. Then, we can evaluate our model using the OSI database as well as social media such as Twitter to determine the level of generalization our model may be able to achieve.

\section{conclusion}
Our paper demonstrates a rapid means of image augmentation and identifying key aspects of protest activity from publicly available image streams, using open source software. Although there are additional work that need to be done to improve our classifier models, our approach creates greater opportunities for the collection of such data to enable work for public good. While traditional efforts to monitor violence and protests may largely be hampered by linguistic barriers and reporting delays, images streams from social media provide a language-agnostic means of assessing such threats. By demonstrating that we were able to get reasonable prediction accuracy of key aspects of protest images using SVM and CNNs, we hope to enable its application to improve monitoring of social unrest activities within unstable regions\cite{sun2021multi}.
\vspace{1em}

\addtolength{\textheight}{-12cm}   

\vspace{-2em}
\end{spacing}
\section*{ACKNOWLEDGMENT}
\vspace{-0.5em}
We thank Virginia Tech CS department for providing us with the OSI dataset. 

\bibliographystyle{unsrtnat}{
\bibliography{main}
}
\end{document}